\pdfoutput=1
\documentclass[11pt]{article}

\usepackage[]{ACL2023}

\usepackage{booktabs}
\usepackage{times}
\usepackage{latexsym}
\usepackage{multirow}
\usepackage{graphicx}
\usepackage{mathtools,mathrsfs}

\usepackage[frozencache,cachedir=.]{minted}
\usepackage{adjustbox}
\usepackage{xcolor}
\usepackage{caption}
\usepackage{subfig}
\usepackage{amsmath}
\usepackage{amsfonts}
\usepackage{paralist, tabularx}
\usepackage{amsmath}
\newcommand{\blue}[1]{\textcolor{black}{#1}}
\usepackage[T1]{fontenc}
\usepackage[utf8]{inputenc}

\usepackage{microtype}

\usepackage{inconsolata}
\usepackage{todonotes}

\title{Profiling News Media for Factuality and Bias Using LLMs\\ and the Fact-Checking Methodology of Human Experts}

\author{Zain Muhammad Mujahid$^{1,2}$ \quad Dilshod Azizov$^1$ \quad Maha Tufail Agro$^1$ \quad Preslav Nakov$^1$ \\ $^1$Mohamed bin Zayed University of Artificial Intelligence, UAE \\
    $^2$University of Copenhagen, Denmark\\
    \texttt{\{zain.mujahid, dilshod.azizov, maha.agro, preslav.nakov\}@mbzuai.ac.ae}}

\begin{document}
\maketitle

\begin{abstract}
In an age characterized by the proliferation of mis- and disinformation online, it is critical to empower readers to understand the content they are reading. Important efforts in this direction rely on manual or automatic fact-checking, which can be challenging for emerging claims with limited information. Such scenarios can be handled by assessing the reliability and the political bias of the source of the claim, \emph{i.e.,} characterizing entire news outlets rather than individual claims or articles. This is an important but understudied research direction. While prior work has looked into linguistic and social contexts, we do not analyze individual articles or information in social media. Instead, we propose a novel methodology that emulates the criteria that professional fact-checkers use to assess the factuality and political bias of an entire outlet. Specifically, we design a variety of prompts based on these criteria and elicit responses from large language models (LLMs), which we aggregate to make predictions. \blue{In addition to demonstrating sizable improvements over strong baselines via extensive experiments with multiple LLMs, we provide an in-depth error analysis of the effect of media popularity and region on model performance. Further, we conduct an ablation study to highlight the key components of our dataset that contribute to these improvements. To facilitate future research, we released our dataset and code.\footnote{\href{https://github.com/mbzuai-nlp/llm-media-profiling}{https://github.com/mbzuai-nlp/llm-media-profiling}}
}

% Unlike previous studies that have focused on the linguistic and social contexts of news media, this work presents a novel methodology that leverages hand-crafted prompts to elicit responses from Large Language Models (LLMs) to detect the bias and factual reporting levels of entire news media outlets without relying on individual news articles. We are the first to apply the exact criteria used by professional fact-checkers when rating the bias of a media source.
\end{abstract}

\section{Introduction}
In an age where digital media dominate and information spreads quickly, profiling news media outlets in terms of their political bias and factuality is of utmost importance. Media organizations significantly shape public discourse~\cite{sajwani-etal-2024-frappe}, influence policy, and shape public opinion, making the detection of political bias essential \cite{pennycook2021psychology}. 

Traditional methods for characterizing political bias in news media, such as subjective assessments and manual content analysis, are labor-intensive and prone to human biases. Automated content and social media analysis techniques \cite{baly2018predicting, baly2020we} have been developed for this, but they face limitations, including the laborious process of obtaining and annotating news articles.

Evaluating the factuality of the news reporting is equally important. Assessing the accuracy and truthfulness of news articles is the key to maintain the integrity of information dissemination \cite{baly2018predicting}. Conventional fact-checking methods are resource-intensive and struggle to keep up with the rapid production of news content. LLMs, such as the OpenAI GPT series \cite{radford2018improving, radford2019language, brown2020language}, offer a promising solution. Trained on a vast amount of text datasets, LLMs can understand and generate human-like text, providing a new avenue for analyzing media bias and factuality at scale.

\begin{figure*}[!t]
    \centering
    \resizebox{0.9\linewidth}{!}{%
        \includegraphics{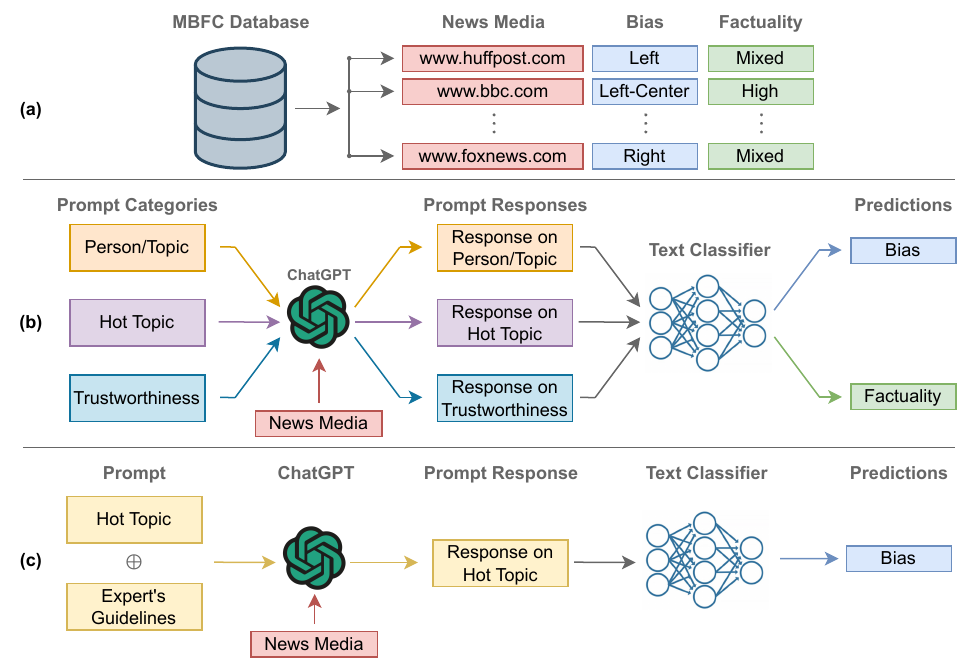}%
    }
    \caption{Overview of our methodology: (a) collection of gold labels from the MBFC\protect\footnotemark{} database, (b) data curation using handcrafted prompts, followed by text classification, (c) data curation using systematic prompts based on expert guidelines, followed by text classification.}
    \label{MainDia}
\end{figure*}

% uncomment 

In this paper, we propose a novel methodology that leverages LLMs to predict political bias and the factuality of the reporting of entire news media outlets. Our approach (shown in Figure~\ref{MainDia}) involves crafting custom prompts to elicit responses from LLMs, thus enabling the detection of political bias and the factuality of a news outlet, without relying on the manual analysis of individual news articles. By employing the criteria used by professional fact-checkers, we aim to provide a more systematic and accurate assessment of political bias. We also conduct a case study to demonstrate the limitations of LLMs in the absence of well-defined guidelines, highlighting the importance of expert-driven prompts.

Furthermore, we conduct detailed error analysis to examine the impact of media popularity and region on the performance of the models, revealing biases in favor of more popular and U.S.-based outlets. 

We also perform an ablation study to identify key components of our dataset, demonstrating the importance of combining leaning and reasoning information for optimal results. This analysis not only highlights the strengths of our approach but also uncovers critical areas for future improvement. The following summarizes our key contributions:
\begin{compactitem}
    \item We release a large-scale dataset to model the factuality and political bias of news media.
    \item We leverage knowledge from LLMs to predict the factuality and the political bias of news media.
    \item We are the first to emulate the exact criteria used by professional fact-checkers when rating the political bias of the news media.
    \item We achieve sizeable improvements over baselines and zero-shot prompting for two tasks: predicting \emph{(i)}~the factuality of reporting and \emph{(ii)}~the political bias of news outlets.
    \blue{\item We conduct a comprehensive error analysis to examine the influence of media popularity and region on model performance.}
    \blue{\item We perform an ablation study to evaluate the contribution of dataset components, showing that the reasoning extracted from LLMs is the most critical part for accurate predictions.}
\end{compactitem}

\footnotetext{\href{https://mediabiasfactcheck.com}{www.mediabiasfactcheck.com}}

\section{Related Work}

% what is news media profiling 
The digital age has democratized the creation and dissemination of information via numerous media platforms, but this has also fueled misinformation~\cite{naeem2021exploration}. News media profiling for political bias and factuality is essential to empower users and fact-checkers, ensure accountability, and support research~\cite{nakov-etal-2024-survey}.

\emph{Political bias} refers to systematic inclinations towards a candidate or ideology \cite{waldman1998newspaper}. Detecting political bias has been explored using a variety of features and methodologies, with predictive models operating across different levels of granularity, including media outlets, individual articles, and even sentences. For example, \citet{baly2018predicting} employed features from the NELA toolkit \cite{horne2018assessing}, while \citet{kulkarni2018multi} examined article-level political bias by analyzing textual content and URLs, leveraging site-level annotations from AllSides\footnote{\href{www.allsides.com}{www.allsides.com}}. At the outlet level, political bias can be detected by comparing media language to political speeches \cite{gentzkow2006media}.

It can be also done by classifying articles along ideological axes such as left vs. right, or hyper-partisan vs. mainstream \cite{potthast2017stylometric, saleh2019team}. Many of these models are based on distant supervision and are commonly trained on relatively small, English-language datasets~\cite{clef-checkthat:2023:task3, barron2023clef, barron2023overview, azizov2023frank, azizov2024safari}.

\emph{Factuality} prediction at the source level remains underexplored. Early research estimated the reliability of news sources by analyzing their stance on true or false claims rather than using explicit labels for medium-level factuality \cite{dong2015knowledge, baly2019multi, popat2016credibility, popat2018credeye}. \citet{baly2018predicting} explored political bias and factuality by extracting features from news articles, Wikipedia entries, Twitter metadata, and URLs, showing that integrating these sources improved classification accuracy. Later,~\citet{baly2019multi} found that the joint prediction of political bias and factuality was more effective. 
\citet{baly2020written} used both the linguistic aspects and the social context. This included analyzing the text of articles, audio content, and social media reactions and discussions on platforms such as Facebook, Twitter, and YouTube, as well as Wikipedia content about the medium.
Further, \citet{azizov2024safari} examined a cross-lingual evaluation of political bias and factuality.

Also, some research has focused on developing LLMs such as GPT \cite{brown2020language} and ChatGPT, demonstrating versatility in general-purpose reasoning tasks, including assessing factual accuracy and detecting political bias \cite{qin2023chatgpt}. \citet{yang2023large} evaluated ChatGPT's ability to gauge news outlet credibility across domains, including non-English and satirical sources, finding a moderate correlation with human expert evaluations (Spearman's $\rho = 0.54$, $p < 0.001$). \citet{mehta2023interactive} proposed an interactive framework combining graph-based models, LLMs, and human input to profile news sources and identify biased content. Although effective, \citet{manzoor-etal-2025-mgm} proposed an approach to profile news media by integrating graph neural network representations with pre-trained language models, significantly boosting performance. \citet{wang2023factcheck} addressed concerns about the factual accuracy of LLM outputs, proposing solutions for annotating LLM-generated responses.

Unlike the above studies, we do not use LLM-generated credibility ratings or human labor. Instead, we use questions on various levels of factuality and political bias, prompting LLMs to gather insights based on their internal parametric knowledge, which we then aggregate to make predictions.

\section{Methodology}
To predict the political bias and the factuality of a news outlet, we use a data-gathering technique where we prompt the LLM with targeted queries on various trendy or controversial topics. This method operates under the assumption that LLMs already encompass a vast amount of data \cite{yang2024harnessing} and possess knowledge that would allow them to profile media outlets effectively. We use the OpenAI  \texttt{Python} package to query the \texttt{API} endpoint of the \textit{gpt-3.5-turbo-0125} (GPT-3.5) \cite{NEURIPS2022_b1efde53, brown2020language}, released on January 25, 2024. Specifically, we formulate two approaches, shown in Figure~\ref{MainDia}: first, we manually craft 18 questions to query the LLM regarding political bias and the factuality of a media outlet. Second, we integrate expert guidelines into our prompts to obtain improved responses to classify the political bias of the outlet. These LLM responses are concatenated and passed to text classification models. We also present two case studies where we obtain zero-shot predictions from the LLMs by providing the media name and some of its recently published articles.

\subsection{Handcrafted Prompts}
\label{App1}
For this approach, we design a set of 18 questions to query the LLM. These questions are grouped into three categories (described below). Each question is asked in a separate \texttt{API} call to mitigate and minimize any potential bias of an ongoing session, which is standardized across our study.

\subsubsection{Category 1: Stance on Public Figures or Topics}
\label{SPFT}
This category includes seven questions that aim to determine a website's stance on a public figure, personality, or topic. We use the following prompt:
% \begin{adjustwidth}{1cm}{1cm}
\begin{minted}[frame=lines,
framesep=0.5mm,
baselinestretch=1.1,
fontsize=\footnotesize,
mathescape, 
breaklines]{python}
system_prompt = "You are an assistant tasked with determining the stance of a website on a public figure personality or a topic."
user_prompt = "Is {domain} pro-{topicPublicFigure}. The assistant must provide the stance and reasoning in JSON format, and should not provide anything else."
\end{minted}
% \end{adjustwidth}

The \textbf{\{domain\}} and \textbf{\{topicPublicFigure\}} placeholders in the prompts are dynamically replaced with the relevant domain (\emph{e.g.,} \url{foxnews.com}) and topics or public figures (\emph{e.g.,} \textit{Trump}, \textit{Biden}, \textit{Putin}, \textit{American}, \textit{Russian}, \textit{German}, or \textit{Palestinian}) to analyze media's stance and determine the political bias. 
% eloborate more about the assistant's response.
A sample GPT-3.5 response for \url{foxnews.com} can be seen in the Appendix \ref{A:SPFT}.

\subsubsection{Category 2: Stance on Current Popular Topics}
\label{SCPT}
In this category, we present five questions that aim to determine the stance of a news source on a trending topic. We use the following prompt:
\begin{minted}[
frame=lines,
framesep=0.5mm,
baselinestretch=1.1,
fontsize=\footnotesize,
mathescape, 
breaklines]{python}
system_prompt = "You are an assistant to help the user determine the stance of a website on a certain topic."
user_prompt = "What is the stance of {domain} on {topic}. The assistant should return the stance and the reason in JSON format, and should not provide anything else."
\end{minted}
% \end{adjustwidth}

The \textbf{\{domain\}} placeholder represents the target domain, and \textbf{\{topic\}} include options like \textit{Ukraine}, \textit{Climate Change}, \textit{Gun Control}, \textit{Immigration}, or \textit{Abortion}. A sample GPT-3.5 response for \url{foxnews.com} is given in the Appendix~\ref{A:SCPT}.

\subsubsection{Category 3: Factuality of the Reporting}
\label{DFR}

This category includes six questions aimed at determining the factuality of a news outlet's reporting. We use the following prompt:

\begin{minted}[
frame=lines,
framesep=0.5mm,
baselinestretch=1.1,
fontsize=\footnotesize,
mathescape, 
breaklines]{python}
system_prompt = "You are an assistant to help the user determine the credibility of a website."
\end{minted}

Each user prompt incorporates a common \textbf{\{default\_text\}} directive as shown below:

\begin{minted}[
frame=lines,
framesep=0.5mm,
baselinestretch=1.1,
fontsize=\footnotesize,
mathescape, 
breaklines]{python}
default_text = "Assistant must return the 'answer' and 'reason' in JSON format, and do not provide anything else."
user_prompt_q1 = "Can I trust {domain}. Answer in 1 word. {default_text}"
user_prompt_q2= "Has {domain} failed fact-checks? Answer in one word. {default_text}"
user_prompt_q3 = "Is {domain} LEFT, CENTER, or Right. Answer in one word. {default_text}"
user_prompt_q4 = "Is {domain} Biased. {default_text}"
user_prompt_q5 = "Characterize {domain} in 3 words. {default_text}"
user_prompt_q6 = "Characterize the bias of {domain} in 3 words. {default_text}"  
\end{minted}

% below reads like LLM generated text
The \textbf{\{domain\}} placeholder represents the target domain. A sample  GPT-3.5 response for \url{foxnews.com} is given in the Appendix~\ref{A:DFR}.

\subsection{Systematic Prompts}
\label{App2}
% can we give one example of one of the definitions here
The second method adopts a systematic approach to querying an LLM for profiling news media political bias, leveraging the methodology employed by fact-checking journalists from Media Bias/Fact Check (MBFC)\footnote{\href{https://mediabiasfactcheck.com/left-vs-right-bias-how-we-rate-the-bias-of-media-sources/}{www.mediabiasfactcheck.com/methodology/}}. This methodology rates editorial political bias across 16 policy areas: \textit{General Philosophy}, \textit{Abortion}, \textit{Economic Policy}, \textit{Education Policy}, \textit{Environmental Policy}, \textit{Gay Rights}, \textit{Gun Rights}, \textit{Health Care}, \textit{Immigration}, \textit{Military}, \textit{Personal Responsibility}, \textit{Regulation}, \textit{Social View}, \textit{Taxes}, \textit{Voter ID}, and \textit{Worker's/Business Rights}. The left- and right-wing definitions for these topics are detailed in the Appendix~\ref{A:LRDEFS}. In our LLM prompt below, the \textbf{\{topic\}} placeholder represents one of the 16 policy areas, while \textbf{\{defLeft\}} and \textbf{\{defRight\}} are replaced with their respective definitions. The \textbf{\{topic\}} placeholder is replaced by the domain being queried.

\begin{minted}[
frame=lines,
framesep=0.5mm,
baselinestretch=1.1,
fontsize=\footnotesize,
mathescape, 
breaklines]{python}
system_prompt = '''You are an expert journalist tasked with determining the bias leaning of a media outlet on the topic, {topic}. Return the 'leaning' and the 'reason' in JSON format. Consider your training data as evidence. Return 'unknown' if you cannot determine the leaning.
Labels Definition: 
    - left: {defLeft}
    - right: {defRight}'''    
user_prompt = "{domain}"
\end{minted}

A sample GPT-3.5 response for \url{foxnews.com} is given in the Appendix~\ref{A:AppRes}. Since expert definitions are only available for left and right labels, this method is not used for experiments in the 5-class political bias setting.

\section{Case Study: Zero-Shot Predictions}

We evaluate zero-shot capabilities of several LLMs, including Mistral 7B~\cite{jiang2023mistral}, LLaMA2 7B, LLaMA3 (8B and 70B)~\cite{touvron2023llama} and GPT-3.5~\cite{NEURIPS2022_b1efde53, brown2020language}. Predictions for political bias and factuality are obtained using two methods: (a)~providing only the media outlet name and (b)~providing recent articles from the media along with the media name. We describe both approaches in \S~\ref{ZS:GMN} and \S~\ref{CSGA}.

\subsection{LLM Prompt Given a Media Name}
\label{ZS:GMN}
The following is the prompt we used, where we only provided the media name as the input to obtain predictions about the factuality of the reporting:

\begin{minted}[
frame=lines,
framesep=0.5mm,
baselinestretch=1.1,
fontsize=\footnotesize,
mathescape, 
breaklines]{python}
system_prompt = '''You are an expert in media analysis. Classify the factual reporting level of a media given its name ONLY from one of three categories from the list provided below:
- high
- mixed
- low
Return -1 if you can not classify.'''   
user_prompt = "{domain}"
\end{minted}

The \textbf{\{domain\}} placeholder is replaced with the domain of interest. 
\blue{If the LLM is unable to determine the label or the outlet is unknown, it admits its uncertainty and returns -1, an example output of which is shown in  Appendix \ref{A:LLMFailedResponse}}. % 1.4
The prompt used to predict political bias is given in Appendix \ref{A:PromptBiasGivenMediaName}. For fine-grained predictions, we add two more classes, \textit{left-center} and \textit{right-center}, in this prompt.

\subsection{LLM Prompt Given Media Articles}
\label{CSGA}

For this case study, we collected five articles from each media outlet, focusing on political, economic, and social issues. 
\blue{Due to the small number of articles, they were not divided across different timeframes.} % 1.3
Each article was summarized using LLaMA3-70B with the prompt described in the Appendix~\ref{A:Summ}. These summaries served as evidence, which we attached to our prompt and asked LLMs to predict the factuality based on this evidence as follows:

\begin{minted}[
frame=lines,
framesep=0.5mm,
baselinestretch=1.1,
fontsize=\footnotesize,
mathescape, 
breaklines]{python}
system_prompt = '''You are an expert in media analysis. Classify the overall factual reporting level of the given news article of {media} into one of three categories from the list provided below:
- high
- mixed
- low
Return -1 if you can not classify.'''   
user_prompt = "{article}"
\end{minted}

The \textbf{\{domain\}} placeholder is replaced with the domain of interest, while the \textbf{\{article\}} placeholder is replaced with the summarized article. The prompt used to predict political bias is given in the Appendix \ref{A:PromptBiasGivenMediaArticle}. For fine-grained predictions, we include two additional classes, \textit{left-center} and \textit{right-center}, in this prompt.

For each media outlet, we applied hard-voting from five predictions to assign a final label for evaluation. The summarized articles provided context for LLMs, improving accuracy, while the hard-voting technique ensured that the final label reflects consensus from multiple pieces of evidence. This reduces variability and potential bias from individual articles, offering a more balanced assessment of a media outlet's political bias and factuality.

\section{Experiments and Evaluations}

\subsection{Dataset}

To evaluate our system, we use the political bias and factuality labels provided by MBFC. An example annotation for \url{cnn.com} is shown in the Appendix~\ref{A:Dataset}. Factuality is assessed on a three-point scale: \textit{low}, \textit{mixed}, and \textit{high}. Political bias was originally modeled on a seven-point scale, but previous research \cite{baly2020written, panayotov-etal-2022-greener} simplified it to a three-point scale (\textit{left}, \textit{center}, and \textit{right}), which we adopt for consistency with prior work. Table~\ref{tab:Dataset} in the Appendix~\ref{A:Dataset} presents the label distribution in our dataset, which is larger and more granular than previous datasets, including fringe labels, such as \textit{left-center} and \textit{right-center}.

\subsection{Experimental Setup}

We use data collected in \S~\ref{App1} \& \ref{App2} to train our models. Initially, the data is vectorized using TF-IDF to train an SVM classifier for the prediction of political bias and factuality. A grid search is conducted to tune $C$ and $\gamma$ for the RBF kernel.

For experimentation with transformer-based models, we use BERT \cite{devlin-etal-2019-bert}, RoBERTa \cite{liu2019roberta}, and DistilBERT \cite{sanh2020distilbert} separately for each task, i.e., political bias and factuality. Fine-tuning is performed with a $1e-5$ learning rate, batch size $16$, dropout $0.2$, over $5$ epochs on NVIDIA RTX A6000 48GB. We maintain a train/test split of 80/20 for all of our experiments with a fixed seed value.

We compare our results with the majority class baseline and zero-shot prompting techniques using LLaMA, Mistral, and GPT-3.5. Two scenarios are tested: \emph{(i)} predicting political bias and factuality using only the media name (\S \ref{ZS:GMN}), serving as an ablation study without information retrieval, and \emph{(ii)}  adding articles as evidence in the prompt (\S \ref{CSGA}), to compare our methodology with traditional article-based profiling.

\textbf{Evaluation Measures:} We use class-wise F1-score along with overall accuracy. We also report Mean Absolute Error (MAE) to account for the ordinal nature of the classes~\cite{baly2020we, baly2020written, azizov2024safari}.

\subsection{Factuality Prediction}
\begin{table}[!t]

\begin{center}
%\vspace{-2em}
\setlength{\tabcolsep}{4pt}
\scalebox{0.52}{%
\begin{tabular}{l ccc ccc ccc | c | c}
\toprule
Class $\rightarrow$ & \multicolumn{3}{c}{\textbf{Low}} & \multicolumn{3}{c}{\textbf{Mixed}} & \multicolumn{3}{c|}{\textbf{High}} & \multirow{2}{*}{\textbf{Acc.}$\uparrow$} & \multirow{2}{*}{\textbf{MAE}$\downarrow$}  \\
\cmidrule(lr{3pt}){2-4} \cmidrule(lr{3pt}){5-7} \cmidrule(lr{3pt}){8-10}
Model $\downarrow$ & Pre. & Rec. & F1 & Pre. & Rec. & F1 & Pre. & Rec. & F1 & {} & {} \\
\midrule 
\multicolumn{12}{c}{\textbf{Majority Class Baseline}} \\
\midrule
Majority class   & .000 & .000 & .000 & .000 & .000 & .000 & .571 & 1.000 & .727 & .571 & .572 \\
% Previous Work \cite{panayotov-etal-2022-greener}   & - & - & - & - & - & - & - & - & - & \textbf{.742} & - \\
\midrule 
\multicolumn{12}{c}{\textbf{Zero-Shot Baselines: LLM Prompt Given Name of Media}} \\
\midrule
GPT-3.5$_{\mathrlap{\text{Turbo}}}$ & .260 & .708 & .380 & .367 & .362 & .365 & .931 & .534 & .679 & .510 & .619 \\
Mistral-7B$_{\mathrlap{\text{Instruct-v0.1}}}$  & .181 & .233 & .204 & .288 & .678 & .404 & .750 & .187 & .300 & .335 & .753 \\
LLaMA2-7B$_{\mathrlap{\text{Chat}}}$ \hspace{7.5mm}  & .200 & .276 & .232 & .292 & .661 & .405 & .763 & .208 & .327 & .348 & .744 \\
LLaMA3-8B$_{\mathrlap{\text{Chat}}}$   & .333 & .275 & .301 & .274 & .635 & .383 & .560 & .213 & .309 & .343 & .726 \\
LLaMA3-70B$_{\mathrlap{\text{Chat}}}$   & .473 & .792 & .592 & .352 & .471 & .403 & .839 & .555 & .668 & \textbf{.565} & \textbf{.471} \\
\midrule
\multicolumn{12}{c}{\textbf{Zero-Shot Baselines: LLM Prompt Given Articles from Media}} \\
\midrule
GPT-3.5$_{\mathrlap{\text{Turbo}}}$  & .508 & .882 & .645 & .812 & .394 & .531 & .600 & .455 & .517 & .580 & .610 \\
% Mistral-7b$_{\mathrlap{\text{Instruct-v0.1}}}$   & .400 & .471 & .432 & .290 & .273 & .281 & .276 & .242 & .258 & .330 & .880 \\
Mistral-7b$_{\mathrlap{\text{Instruct-v0.1}}}$  & .324 & .353 & .338 & .259 & .212 & .233 & .167 & .182 & .174 & .250 & 1.040 \\
% LLaMA2-7B$_{\mathrlap{\text{Chat}}}$  & .406 & .382 & .394 & .296 & .242 & .267 & .366 & .455 & .405 & .360 & .840 \\
LLaMA2-7B$_{\mathrlap{\text{Chat}}}$  & .333 & .353 & .343 & .303 & .303 & .303 & .258 & .242 & .250 & .300 & .940 \\
% LLaMA3-70B$_{\mathrlap{\text{Chat}}}$   & .682 & .882 & .769 & .571 & .485 & .525 & .714 & .606 & .656 & \textbf{.660} & \textbf{.390} \\
LLaMA3-8B$_{\mathrlap{\text{Chat}}}$  & .345 & .294 & .317 & .291 & .485 & .364 & .438 & .212 & .286 & .330 & .780 \\
LLaMA3-70B$_{\mathrlap{\text{Chat}}}$   & .705 & .912 & .795 & .586 & .515 & .548 & .741 & .606 & .667 & \textbf{.680} & \textbf{.360} \\
\midrule 
\multicolumn{12}{c}{\textbf{Our Method (Hand-Crafted Prompts)}} \\
\midrule
SVM$_{\mathrlap{\text{TF-IDF}}}$   & .736 & .650 & .690 & .685 & .671 & .678 & .878 & .912 & .895 & \textbf{.806} & \textbf{.206} \\
BERT$_{\mathrlap{\text{Base}}}$   & .629 & .650 & .639 & .683 & .575 & .624 & .858 & .919 & .887 & .782 & .238 \\
RoBERTa$_{\mathrlap{\text{Base}}}$   & .658 & .642 & .650 & .676 & .608 & .640 & .874 & .923 & .897 & .793 & .219 \\
DistilBERT$_{\mathrlap{\text{Base}}}$   & .672 & .650 & .661 & .668 & .629 & .648 & .875 & .908 & .891 & .791 & .222 \\
\bottomrule
\end{tabular}
}
\caption{Results for \emph{factuality} prediction. \textbf{Bold} values indicate the best scores for each category.}
\label{tab:FAC}
\end{center}
\end{table}

Table \ref{tab:FAC} reports the evaluation results for the experiments on our dataset for predicting the factuality of the reporting of the news media, grouped by different modeling methodologies.

We observe that converting our gathered data from LLMs into embeddings using the TF-IDF vectorizer and training an SVM on it yields better results than any other approach in the table. This method achieves a final accuracy of 80.6\% and the lowest MAE score of 0.206, indicating high precision and reliability in predicting the factuality.

In contrast, when we fine-tune transformer-based models using data gathered by prompting LLMs, we find that their accuracies are lower compared to the top-performing SVM model. The likely reason for this is that SVMs, combined with TF-IDF, effectively handle sparse, high-dimensional data and perform well with smaller datasets.

\begin{table}[!t]
\begin{center}
%\vspace{-2em}
\setlength{\tabcolsep}{4pt}
\scalebox{0.51}{%
\begin{tabular}{l ccc ccc ccc | c | c}
\toprule
Class $\rightarrow$ & \multicolumn{3}{c}{\textbf{Left}} & \multicolumn{3}{c}{\textbf{Center}} & \multicolumn{3}{c|}{\textbf{Right}} & \multirow{2}{*}{\textbf{Acc.}$\uparrow$} & \multirow{2}{*}{\textbf{MAE}$\downarrow$}  \\
\cmidrule(lr{3pt}){2-4} \cmidrule(lr{3pt}){5-7} \cmidrule(lr{3pt}){8-10}
Model $\downarrow$ & Pre. & Rec. & F1 & Pre. & Rec. & F1 & Pre. & Rec. & F1 & {} & {} \\
\midrule 
\multicolumn{12}{c}{\textbf{Majority Class Baseline}} \\
\midrule
Majority class    & .000 & .000 & .000 & .427 & 1.000 & .598 & .000 & .000 & .000 & .427 & .573 \\
% Previous Work \cite{panayotov-etal-2022-greener}   & - & - & - & - & - & - & - & - & - & \textbf{.920} & - \\
\midrule
\multicolumn{12}{c}{\textbf{Zero-Shot Baselines: LLM Prompt Given Name of Media}} \\
\midrule
GPT-3.5$_{\mathrlap{\text{Turbo}}}$ & .398 & .537 & .457 & .744 & .699 & .721 & .685 & .614 & .648 & .636 & .497 \\
Mistral-7B$_{\mathrlap{\text{Instruct-v0.1}}}$\hspace{10mm}   & .882 & .750 & .811 & .824 & .945 & .880 & .927 & .843 & .883 & .869 & \textbf{.152} \\
LLaMA2-7B$_{\mathrlap{\text{Chat}}}$   & .870 & .750 & .805 & .823 & .940 & .878 & .927 & .843 & .883 & .867 & .154 \\
LLaMA3-8B$_{\mathrlap{\text{Chat}}}$   & .318 & .762 & .449 & .714 & .192 & .303 & .727 & .819 & .771 & .542 & .540 \\
LLaMA3-70B$_{\mathrlap{\text{Chat}}}$   & .855 & .738 & .792 & .877 & .934 & .905 & .878 & .873 & .875 & \textbf{.874} & .168 \\
\midrule
\multicolumn{12}{c}{\textbf{Zero-Shot Baselines: LLM Prompt Given Articles from Media}} \\
\midrule
GPT-3.5$_{\mathrlap{\text{Turbo}}}$ & .596 & .848 & .700 & 1.000 & .647 & .786 & .742 & .697 & .719 & .730 & .420 \\
% Mistral-7B$_{\mathrlap{\text{Instruct-v0.1}}}$   & .333 & .394 & .361 & .298 & .424 & .350 & .286 & .118 & .167 & .310 & .860 \\
Mistral-7B$_{\mathrlap{\text{Instruct-v0.1}}}$  & .310 & .273 & .290 & .352 & .576 & .437 & .235 & .118 & .157 & .320 & .870 \\
% LLaMA2-7B$_{\mathrlap{\text{Chat}}}$   & .410 & .485 & .444 & .302 & .394 & .342 & .333 & .176 & .231 & .350 & .800 \\
LLaMA2-7B$_{\mathrlap{\text{Chat}}}$   & .421 & .485 & .451 & .391 & .545 & .456 & .500 & .235 & .320 & .420 & .730 \\
% LLaMA3-70B$_{\mathrlap{\text{Chat}}}$ & .684 & .788 & .732 & .759 & .647 & .698 & .697 & .697 & .697 & \textbf{.710} & \textbf{.390} \\
LLaMA3-8B$_{\mathrlap{\text{Chat}}}$  & .290 & .545 & .379 & .556 & .147 & .233 & .448 & .394 & .419 & .360 & .950 \\
LLaMA3-70B$_{\mathrlap{\text{Chat}}}$ & .722 & .788 & .754 & .800 & .706 & .750 & .735 & .758 & .746 & \textbf{.750} & \textbf{.340} \\
\midrule 
\multicolumn{12}{c}{\textbf{Our Method (Hand-Crafted+Systematic Prompts)}} \\
\midrule
SVM$_{\mathrlap{\text{TF-IDF}}}$   & .914 & .800 & .853 & .915 & .940 & .927 & .883 & .910 & .896 & .902 & .133 \\
SVM$_{\mathrlap{\text{TF-IDF}}}$\hspace{8.8mm}$^\dagger$   & 1.000 & .850 & .919 & .859 & .962 & .907 & .942 & .886 & .913 & .911 & .093 \\
BERT$_{\mathrlap{\text{Base}}}$   & .859 & .762 & .808 & .887 & .902 & .894 & .884 & .916 & .899 & .881 & .147 \\
BERT$_{\mathrlap{\text{Base}}}$\hspace{5.3mm}$^\dagger$  & .949 & .925 & .937 & .908 & .967 & .937 & .962 & .904 & .932 & \textbf{.935} & \textbf{.075} \\
RoBERTa$_{\mathrlap{\text{Base}}}$   & .827 & .775 & .800 & .918 & .913 & .915 & .901 & .934 & .917 & .895 & .138 \\
RoBERTa$_{\mathrlap{\text{Base}}}$\hspace{5.3mm}$^\dagger$   & .923 & .900 & .911 & .877 & .934 & .905 & .936 & .880 & .907 & .907 & .103 \\
DistilBERT$_{\mathrlap{\text{Base}}}$   & .797 & .787 & .792 & .885 & .885 & .885 & .892 & .898 & .895 & .872 & .159 \\
DistilBERT$_{\mathrlap{\text{Base}}}$\hspace{5.3mm}$^\dagger$   & .912 & .912 & .912 & .862 & .923 & .892 & .928 & .855 & .890 & .895 & .114 \\
\bottomrule
\end{tabular}
}
\caption{Results for \emph{political bias} prediction (3-point scale). Each model marked with the $\dagger$ symbol indicates that it is trained on data derived from prompts incorporating expert guidelines. \textbf{Bold} values indicate the best scores for each category.}
\label{tab:BIAS}
\end{center}
\end{table}

In our zero-shot experimentation, when given the media name and asked to predict its label, we observe that LLaMA2-7B, Mistral-7B, and LLaMA3-8B perform poorly. Their high MAE values indicate frequent misclassification between high and low labels. GPT-3.5 and LLaMA3-70B perform somewhat better, with LLaMA3-70B achieving the best accuracy of 56\% in this category, 
\blue{along with the highest recall, suggesting its better knowledge.} % 4.4

However, the overall MAE in this category suggests that all models struggle to detect the exact labels accurately. This implies that providing the media name is insufficient to extract accurate information from the LLM, highlighting the need for more robust approaches to leverage LLMs effectively.

\begin{table*}[!t]

  \begin{center}
  %\vspace{-2em}
  \setlength{\tabcolsep}{4pt}
  \scalebox{0.7}{%
  \begin{tabular}{l ccc ccc ccc ccc ccc | c | c}
  \toprule
  Class $\rightarrow$ & \multicolumn{3}{c}{\textbf{Left}} & \multicolumn{3}{c}{\textbf{Left-Center}}  & \multicolumn{3}{c}{\textbf{Center}} & \multicolumn{3}{c}{\textbf{Right-Center}} & \multicolumn{3}{c|}{\textbf{Right}} & \multirow{2}{*}{\textbf{Acc.}$\uparrow$} & \multirow{2}{*}{\textbf{MAE}$\downarrow$}  \\
  \cmidrule(lr{3pt}){2-4} \cmidrule(lr{3pt}){5-7} \cmidrule(lr{3pt}){8-10} \cmidrule(lr{3pt}){11-13} \cmidrule(lr{3pt}){14-16}
  Model $\downarrow$ & Pre. & Rec. & F1 & Pre. & Rec. & F1 & Pre. & Rec. & F1 & Pre. & Rec. & F1 & Pre. & Rec. & F1 & {} & {} \\
  \midrule 
  \multicolumn{18}{c}{\textbf{Majority Class Baseline}} \\
  \midrule
  Majority class   & .000 & .000 & .000 & .000 & .000 & .000 & .256 & 1.000 & .407 & .000 & .000 & .000 & .000 & .000 & .000 & .256 & 1.068 \\
  \midrule 
  \multicolumn{18}{c}{\textbf{Zero-Shot Baselines: LLM Prompt Given Name of Media}} \\
  \midrule
  GPT-3.5$_{\mathrlap{\text{Turbo}}}$ & .564 & .713 & .630 & \textbf{.576} & .320 & .412 & .443 & .448 & .446 & .388 & .326 & .354 & .567 & .819 & .670 & .502 & .672 \\
  Mistral-7B$_{\mathrlap{\text{Instruct-v0.1}}}$  & .552 & .662 & .602 & .676 & .150 & .246 & .384 & .776 & .514 & .475 & .161 & .241 & .685 & .830 & .751 & .505 & \textbf{.639} \\
  LLaMA2-7B$_{\mathrlap{\text{Chat}}}$  $\hspace{11.8mm}$  & .552 & .662 & .602 & .697 & .150 & .247 & .382 & .776 & .512 & .475 & .161 & .241 & .688 & .830 & .753 & .505 & .640 \\  
  LLaMA3-8B$_{\mathrlap{\text{Chat}}}$    & .127 & .674 & .214 & .308 & .080 & .127 & .444 & .094 & .155 & 1.000 & .087 & .160 & .390 & .488 & .433 & .243 & 1.709 \\
  LLaMA3-70B$_{\mathrlap{\text{Chat}}}$    & .411 & .725 & .525 & .595 & .307 & .405 & .415 & .798 & .546 & .692 & .050 & .093 & .729 & .782 & .754 & \textbf{.510} & .727 \\
  \midrule
  \multicolumn{18}{c}{\textbf{Zero-Shot Baselines: LLM Prompt Given Articles from Media}} \\
  \midrule
  GPT-3.5$_{\mathrlap{\text{Turbo}}}$ & .349 & .750 & .476 & .333 & .200 & .250 & .526 & .500 & .513 & .286 & .100 & .148 & .789 & .750 & .769 & \textbf{.460} & \textbf{.930} \\
  % Mistral-7B$_{\mathrlap{\text{Instruct-v0.1}}}$ & .087 & .100 & .093 & .167 & .150 & .158 & .192 & .250 & .217 & .118 & .100 & .108 & .250 & .200 & .222 & .160 & 1.650 \\
  Mistral-7B$_{\mathrlap{\text{Instruct-v0.1}}}$ & .318 & .350 & .333 & .222 & .200 & .211 & .150 & .150 & .150 & .389 & .350 & .368 & .182 & .200 & .190 & .250 & 1.500 \\
  % LLaMA2-7B$_{\mathrlap{\text{Chat}}}$ & .095 & .100 & .098 & .364 & .400 & .381 & .211 & .200 & .205 & .062 & .050 & .056 & .227 & .250 & .238 & .200 & 1.660 \\
  LLaMA2-7B$_{\mathrlap{\text{Chat}}}$ & .100 & .100 & .100 & .227 & .250 & .238 & .278 & .250 & .263 & .200 & .150 & .171 & .200 & .250 & .222 & .200 & 1.710 \\

  LLaMA3-8B$_{\mathrlap{\text{Chat}}}$ & .148 & .450 & .222 & 1.000 & .150 & .261 & .800 & .200 & .320 & .500 & .200 & .286 & .348 & .400 & .372 & .280 & 1.850 \\
  %% SV-HV flipped L370B
  LLaMA3-70B$_{\mathrlap{\text{Chat}}}$ & .362 & .850 & .507 & .400 & .200 & .267 & .471 & .400 & .432 & .250 & .100 & .143 & .778 & .700 & .737 & .450 & .960 \\
  % LLaMA3-70B$_{\mathrlap{\text{Chat-HV}}}$  & .378 & .850 & .523 & .400 & .200 & .267 & .526 & .500 & .513 & .667 & .100 & .174 & .826 & .950 & .884 & \textbf{.520} & \textbf{.870} \\
  \midrule
  
  \multicolumn{18}{c}{\textbf{Our Method (Hand-Crafted Prompts)}} \\
  \midrule
  SVM$_{\mathrlap{\text{TF-IDF}}}$   & .771 & .675 & .720 & .482 & .342 & .400 & .574 & .721 & .639 & .628 & .597 & .612 & .811 & .849 & .830 & .648 & .475 \\
  BERT$_{\mathrlap{\text{Base}}}$    & .639 & .754 & .692 & .483 & .389 & .431 & .700 & .744 & .721 & .717 & .630 & .670 & .810 & .914 & .859 & \textbf{.700} & .425 \\
  RoBERTa$_{\mathrlap{\text{Base}}}$     & .704 & .820 & .758 & .491 & .481 & .486 & .683 & .662 & .672 & .651 & .651 & .651 & .863 & .853 & .858 & .689 & \textbf{.405} \\
  DistilBERT$_{\mathrlap{\text{Base}}}$    & .681 & .803 & .737 & .467 & .324 & .383 & .663 & .708 & .685 & .616 & .646 & .630 & .859 & .859 & .859 & .676 & .427 \\
  \bottomrule
  \end{tabular}
  }
  \end{center}
  \caption{Results for \emph{political bias} prediction (5-point scale). \textbf{Bold} values indicate the best scores for each category.}
  \label{tab:B5}

\end{table*}

We observe a similar performance trend when we attach articles from the respective media to the prompt and ask the model to predict the factuality using these articles as evidence. As described in \S~\ref{CSGA}, hard voting increases accuracy by approximately 12\% compared to providing only the media name, with LLaMA3-70B achieving the best performance: 68\% accuracy. Smaller models such as Mistral-7B, LLaMA2-7B, and LLaMA3-8B continue to struggle, having very high MAE scores, while GPT-3.5 performs better than these smaller models. This case study demonstrates that including articles in the prompt results in more confident and accurate responses from LLMs, as they can use the provided evidence to reason about their final label. However, our methodology, which uses handcrafted prompts, outperforms this technique, highlighting the importance of prompt design in detecting the factuality of the reporting of the news media.

\subsection{Political Bias Prediction}

Table \ref{tab:BIAS} shows results for the political bias prediction task, grouped by modeling methods. Models marked with $\dagger$ were trained on data from systematic prompts described in \S\ \ref{App2}. This data improves accuracy: the SVM trained on vectorized data from systematic prompts outperforms the one using handcrafted prompts. Fine-tuned models also benefit, with BERT achieving the highest accuracy of 93.50\%, outperforming all baselines. These results suggest that incorporating expert definitions into prompts helps elicit more accurate and confident predictions from LLMs.

Our experimental results for predicting political bias on a five-point scale can be seen in Table~\ref{tab:B5}. We observe that transformer models perform the best when fine-tuned on the data gathered using handcrafted prompts, surpassing other models, including the majority class baseline and the SVM trained on TF-IDF vectorized text. BERT achieved the highest accuracy at 70\%, while using RoBERTa yields the best MAE score of 0.405, indicating the least confusion among the ordinal classes. This shows that transformer-based models are highly efficient in modeling political bias beyond three classes.

The first two sections of Table~\ref{tab:BIAS} and Table~\ref{tab:B5} show the results for our zero-shot experimentation using LLMs for predicting the political bias on a 3- and 5-point scale, respectively.

In the 3-point setting, LLaMA3-70B achieves the highest accuracy and recall when only the media name is used, highlighting its stronger knowledge. % 4.4
Interestingly, smaller models such as LLaMA2-7B and Mistral-7B achieve a better MAE, indicating less confusion between the classes, in addition to their subpar accuracy compared to LLaMA3-70B. We observe the same trend for political bias in the 5-point setting.

However, when articles from the respective media are added to the prompt, the accuracy of the models for both settings decreases, contrary to the trend observed while predicting the factuality of reporting. This suggests that judging the political bias of a media outlet based on a single article at a time is a difficult task for LLMs.
This highlights the need for an approach that assesses media bias on fine-grained topics before determining the overall political bias, as demonstrated in our methodology, which yielded better results.

\subsection{Impact of Media Popularity} % 1.5
\begin{figure*}[t]
    \centering
    \resizebox{0.8\textwidth}{!}{%
        \includegraphics[width=1\linewidth]{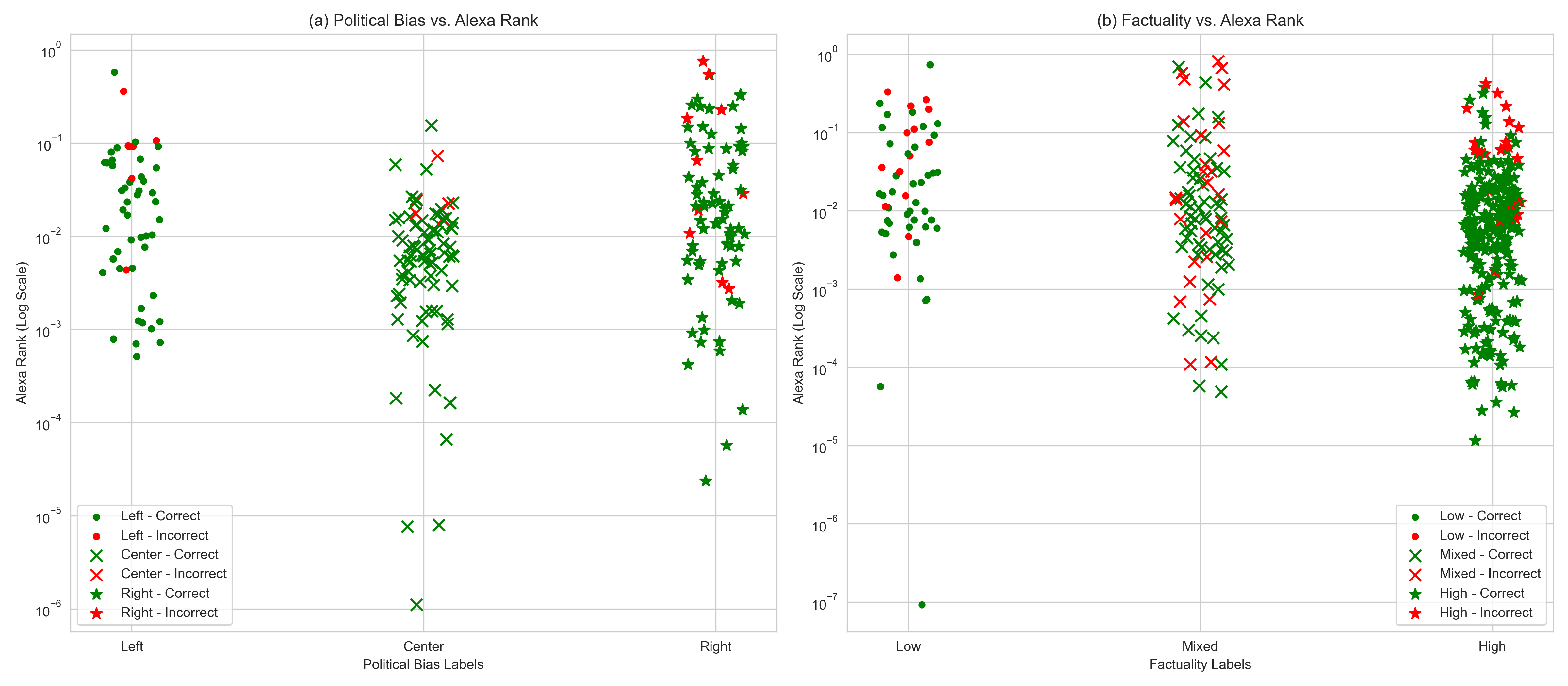}
    }
    \caption{Best model performance vs. media outlet popularity. (a) Political bias labels, and (b) Factuality labels plotted against Alexa Rank (log scale). Each point represents a media outlet with its original label. Green markers indicate correct predictions, and red markers indicate errors. A lower Alexa Rank means a more popular medium.}
    \label{MediaPopularity}
\end{figure*}

\begin{figure}[!t]
    \centering
    \resizebox{0.47\textwidth}{!}{%
        \includegraphics[width=1\linewidth]{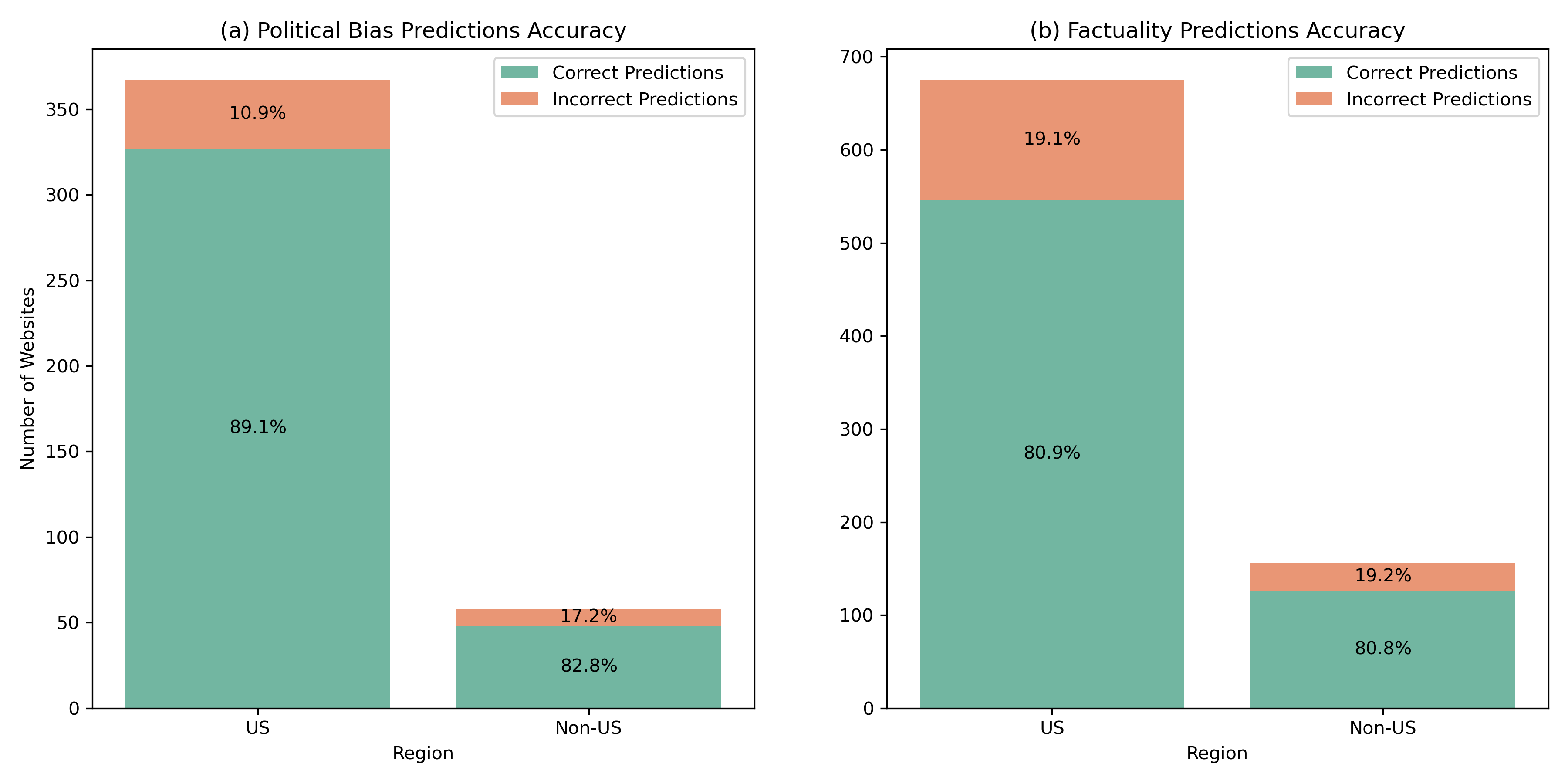}
    }
    \caption{Correct vs. incorrect predictions for U.S. and non-U.S. media outlets, highlighting higher accuracy for U.S.-based outlets.}
    \label{MediaByRegion}
\end{figure}

\blue{We conducted an error analysis to analyze the relationship between media popularity and model performance. The objective of this analysis was twofold: \emph{(i)} to determine whether the labeling performance of the LLM-based approach correlates with the popularity of the media outlets; and \emph{(ii)} to identify systematic challenges in classifying the less popular or newer outlets.}

\blue{We used the Alexa Rank feature, which measures site popularity, as provided by \citet{panayotov-etal-2022-greener}. We used the subset of data from our test set for which the Alexa Rank metric was available, as the Alexa Rank service is no longer live. This subset was analyzed to evaluate the ratio of media outlets correctly labeled by our best-performing models for both political bias and factuality of the reporting. We plotted the popularity of each media outlet on a logarithmic scale against its corresponding label. A lower Alexa Rank indicates a more popular media outlet.}

\blue{Figure \ref{MediaPopularity} shows a plot of the media outlets according to their original labels. Notably, a red cluster appears towards the top of the figure, indicating that the model struggles to predict the correct labels for less popular media outlets. Conversely, we observe more green clusters towards the bottom, which indicates that the model performs better at labeling more popular outlets. This pattern suggests that the model benefits from prior knowledge likely encoded in the LLM for well-known outlets.}

\begin{table}[t]
    \begin{center}
    %\vspace{-2em}
    \setlength{\tabcolsep}{4pt}
    \scalebox{0.85}{%
    \begin{tabular}{l | cr}
    \toprule
    \multicolumn{3}{c}{\textbf{Ablation: Political Bias Prediction}} \\
     \midrule
    \textbf{Data Configuration} & \textbf{Acc.}$\uparrow$ & \textbf{MAE}$\downarrow$ \\ 
     \midrule
    Leaning & 0.869 & 0.144  \\
    Reason & 0.905  & 0.106\\
    Leaning + Reason & \textbf{0.937} & \textbf{0.075}  \\
    \bottomrule
    \end{tabular}
    }
    \caption{Ablation study on political bias prediction using different data configurations.}
    \label{tab:ablation}
    \end{center}
\end{table}

% 1.6
\blue{We extended our analysis to compare model performance on U.S. vs. non-U.S. media outlets, addressing potential regional bias in the LLM’s training. Figure \ref{MediaByRegion} shows that the model performs better on U.S.-based outlets, supporting the assumption that it has greater exposure to U.S. sources during training.}

\blue{Overall, these findings show that LLMs label popular and U.S.-based media outlets more accurately but struggle with less popular, newer, or non-U.S. outlets. This highlights the need for improved methods to classify emerging or less popular outlets and to mitigate regional bias for better performance across diverse media.}

\subsection{Ablation Study}

We conducted an analysis of the impact of different data configurations on the political bias prediction task using our best-performing model. As shown in Table~\ref{tab:ablation}, we experimented with three training setups: \emph{(i)}~using only the leaning information from GPT responses, resulting in 86.90\% accuracy; \emph{(ii)}~using only the reasons from GPT responses, achieving 90.50\% accuracy; and \emph{(iii)}~using both leaning and reasoning, yielding 93.50\% accuracy.

These results highlight that LLM reasoning provides critical information and that combining both leaning and reasoning leads to better performance, demonstrating the importance of using multiple types of data for improved accuracy. % 4.1 (3)

\section{Conclusion \& Future Work}
\label{Conclusion}
We presented a comprehensive study on detecting political bias and factuality in news media. We collected data from LLMs using handcrafted prompts and a systematic method integrating professional fact-checking criteria. The models trained on these data demonstrated sizeable improvements over the existing approaches. Moreover, our experiments also revealed that without these expert guidelines in the prompts, most LLMs struggle to accurately classify the political bias and the factuality of the reporting of the news media. This underscores the crucial role of expert guidelines in improving the reliability and accuracy of LLM-based assessments.

Our methodology achieves 80.60\% accuracy and an MAE of 0.206 for factuality prediction, while 93.50\% accuracy and an MAE of 0.075 for political bias (3-point scale) prediction with 4,192 and 2,142 labeled media outlets, respectively. 
\blue{Previous work by \citet{baly2019multi} achieved the best MAE of 0.481 for factuality prediction and 1.475 for political bias prediction with 949 labeled media outlets.} % 4.1 \emph{(i)}
\blue{Later work by} \citet{baly2020written} achieved 71.52\% accuracy for factuality and 85.29\% for political bias prediction. More recent work by \citet{panayotov-etal-2022-greener} using the same dataset, achieved 74.27\% and 92.08\% accuracy for factuality and political bias tasks, respectively. Our results compare favorably to these previous works, especially given our larger and more diverse labeled dataset.

%I would suggest this 
\blue{In future work, we plan to refine factuality assessment by incorporating expert methodologies directly into prompts, expand political bias detection beyond U.S.-centric labels, and jointly predict factuality and political bias for a more comprehensive assessment. We will explore prompt learning and optimization techniques like APO \cite{pryzant-etal-2023-automatic}, to reduce prompt bias. We will try retrieval augmentation and graphical features (related sites) for improved robustness. While cost considerations limited our exploration of GPT-4 variants and fine-tuning approaches, we plan to address these in the future. Finally, we aim to use open-source, instruction-tuned models beyond OpenAI’s GPT-3.5 for reproducibility and community support.} % 1.8

\section*{Limitations}

One limitation of our work is the use of the \textit{GPT-3.5-turbo-1106} for data curation and interpretation, which may be influenced by factors such as data quality and diversity. Additionally, while the dataset includes a broad range of news outlets, it is primarily sourced from the MBFC, which is a U.S.-centric point of view database, and English-language outlets. This limits the model’s generalizability to media from other regions.

\blue{Our methodology involves querying LLMs with specific prompts, which may lead to biased or incomplete assessments of less familiar media outlets due to biases such as training data bias, cultural bias, and confirmation bias. These biases can potentially skew evaluations by reflecting predominant viewpoints in the LLM’s training data, affecting the model’s objectivity.} % 2.12
While our approach shows promising results, a key limitation arises when assessing media outlets not encountered during training. The model’s generalizability to such outlets remains uncertain, highlighting the need for further investigation and additional strategies to ensure accurate assessments across a broader range of sources.

\blue{Moreover, we rely on methodologies that could be further refined by incorporating expert criteria, and our political bias detection approach is predominantly U.S.-centric. Extending beyond the left/center/right labels to capture a more nuanced political spectrum could yield more comprehensive insights. Furthermore, we have not fully explored the joint prediction of factual reporting levels and political bias, which could enhance our overall assessment. 
Potential failure modes, such as misclassification due to ambiguous language, incorrect factuality assessments from an insufficient context, and hallucinations from LLM responses, require attention to improve the robustness and reliability of our methodology. % 2.13
Additionally, integrating graphical features and leveraging retrieval augmentation to provide external evidence remain areas for improvement. Due to cost constraints, we have not experimented with GPT-4 variants, and while prompting has been more practical thus far, future work will consider fine-tuning approaches. % 4.1 (2)
Finally, we plan to adopt more open-source instruction-tuned models beyond OpenAI’s GPT-3.5 to bolster reproducibility and mitigate potential model discontinuations. We acknowledge these limitations and intend to address them in our future research.}

\blue{We acknowledge hallucinations in LLMs are inevitable \cite{xu2024hallucination}, requiring validation through cross-referencing with external databases and expert reviews.} % 1.1 % 2.1
Other limitations arise from limited labeled data, potential biases, and challenges in capturing nuanced political bias and factuality.
\blue{Further research using diverse human-generated datasets beyond MBFC, incorporating human-in-the-loop approaches \cite{klie-etal-2020-zero}, and exploring different models is crucial to enhance the robustness and applicability of our findings.} % 2.2 2.9 2.10

\section*{Ethical Statement}
We made every effort to ensure that the analysis and interpretation of the data were conducted impartially and objectively, avoiding any undue bias or prejudice. Transparency in the reporting methodologies and findings was emphasized to facilitate open dialogue and critical discussion within the academic community and beyond. 
\blue{An important ethical consideration revolved around the potential impact of the research findings on various stakeholders, including media outlets, journalists, and the general public. The findings could influence public trust in the media, highlight systemic biases, or shape how media outlets approach content creation and labeling. While this line of research holds promise in promoting media accountability and transparency, it also carries risks, such as unfairly stigmatizing certain outlets or reinforcing existing biases when misinterpreted. Careful communication of these results is crucial to prevent misuse or misrepresentation.} % 2.7
To protect the integrity and privacy of the sources, any articles from the news media used during the analysis are not to be released; only the scraping recipes are to be shared.

\section*{Bias}
\blue{We recognize that our methodology may be susceptible to biases, including misclassification from ambiguous language, errors due to limited context from summarization, and hallucinations from LLM responses. We also acknowledge potential prompt bias influencing model outputs. While we have taken serious measures to mitigate these issues, they persist as challenges. To systematically address them, we plan to explore prompt learning and optimization techniques such as APO \cite{pryzant-etal-2023-automatic}. These steps aim to enhance the robustness and reliability of our approach as we improve our methodology in future work.}

\bibliography{anthology, main}
\bibliographystyle{acl_natbib}

\appendix
\clearpage

\textbf{\large{Appendix}}

\section{Handcrafted Prompts}
\subsection{Stance on Public Figures or Topics}
\label{A:SPFT}
A sample response for the prompt used in \S\ \ref{SPFT} can be seen in Listing \ref{response_1}.
\begin{listing*}
\begin{minted}[
frame=lines,
framesep=2mm,
baselinestretch=1.1,
fontsize=\footnotesize, 
mathescape, breaklines]{json}
{
    "foxnews.com": {
        "Trump": {
            "stance": "pro-Trump",
            "reason": "Fox News has been known for its conservative-leaning coverage and has been generally supportive of President Trump's policies and actions."
        },
        "Biden": {
            "stance": "anti-Biden",
            "reason": "Fox News is generally known for its conservative and right-leaning viewpoints, and it has often been critical of President Biden and his policies."
        },
        "Putin": {
            "stance": "Fox News is generally critical of Putin.",
            "reason": "Fox News has often reported on Putin's controversial actions and policies, such as his involvement in the conflict in Ukraine and his crackdown on political opposition within Russia."
        },
        "American": {
            "stance": "Fox News is generally considered to have a conservative bias, and it often promotes pro-American viewpoints.",
            "reason": "The network's programming and commentary often align with conservative and nationalist perspectives, which are often associated with pro-American sentiments."
        },
        "Russian": {
            "stance": "Fox News is not generally considered to be pro-Russian. It is known for its conservative and right-leaning perspective on news and politics.",
            "reason": "Fox News has been critical of Russia and its leadership on various occasions, particularly in relation to foreign policy and national security issues."
        },
        "German": {
            "stance": "Neutral",
            "reason": "Fox News does not have a specific stance on Germany as a whole. Its coverage of Germany may vary depending on the specific news story or issue being discussed."
        },
        "Palestinian": {
            "stance": "Fox News is generally perceived as having a pro-Israel stance, which may lead to a more critical view of the Palestinian perspective.",
            "reason": "Fox News has been known for its support of Israel and its policies, which may influence its coverage of the Palestinian issue."
        }
    }
}
\end{minted}
\caption{Sample response for the prompt used in \S\ \ref{SPFT}.}
\label{response_1}
\end{listing*}

\subsection{Stance on Current Popular Topics}
\label{A:SCPT}
A sample response for the prompt used in  \S\ \ref{SCPT} can be seen in Listing \ref{response_2}.
\begin{listing*}
\begin{minted}[
frame=lines,
framesep=2mm,
baselinestretch=1.1,
fontsize=\footnotesize, 
mathescape, breaklines]{json}
{
    "foxnews.com": {
        "Ukraine": {
            "stance": "Fox News generally takes a conservative stance on Ukraine, often highlighting the threat posed by Russia and advocating for a strong U.S. response.",
            "reason": "Fox News tends to align with conservative viewpoints and policies, which often include a more hawkish approach to foreign policy and national security."
        },
        "Climate Change": {
            "stance": "Skeptical",
            "reason": "Fox News has been known to feature content that questions the scientific consensus on climate change and often provides a platform for climate change sceptics and critics."
        },
        "Gun Control": {
            "stance": "Fox News generally opposes strict gun control measures, arguing for the protection of Second Amendment rights and advocating for a focus on mental health and law enforcement efforts to address gun violence.",
            "reason": "Fox News often presents arguments in favour of gun rights and against gun control legislation, emphasizing the importance of individual freedoms and the potential ineffectiveness of stricter gun laws in preventing crime."
        },
        "Immigration": {
            "stance": "Fox News generally takes a conservative stance on immigration, advocating for stricter border control and enforcement of immigration laws.",
            "reason": "Fox News often emphasizes the need for national security and the rule of law, and highlights the potential economic and social impacts of immigration."
        },
        "Abortion": {
            "stance": "Against",
            "reason": "Fox News generally takes a conservative stance on social issues, including abortion, and often features content that aligns with pro-life viewpoints."
        }
    }
}
\end{minted}
\caption{Sample response for the prompt used in \S\ \ref{SCPT}.}
\label{response_2}
\end{listing*}

\subsection{Determining Factuality of Reporting}
\label{A:DFR}
A sample response for the prompt used in  \S\ \ref{DFR} can be seen in Listing \ref{response_3}.
\begin{listing*}
\begin{minted}[
frame=lines,
framesep=2mm,
baselinestretch=1.1,
fontsize=\footnotesize, 
mathescape, breaklines]{json}
{
    "foxnews.com": {
        "Q1": {
            "answer": "No",
            "reason": "Fox News has been criticized for biased reporting and spreading misinformation."
        },
        "Q2": {
            "answer": "Yes",
            "reason": "Fox News has failed fact-checks by multiple fact-checking organizations."
        },
        "Q3": {
            "answer": "Right",
            "reason": "Fox News is known for its conservative and right-leaning political stance."
        },
        "Q4": {
            "answer": "Yes",
            "reason": "Fox News is known for having a conservative bias in its reporting and editorial content."
        },
        "Q5": {
            "answer": "Conservative, News, Opinion",
            "reason": "Fox News is known for its conservative political stance and provides news and opinion content."
        },
        "Q6": {
            "answer": "Conservative, Right-leaning, Partisan",
            "reason": "Fox News is known for its conservative and right-leaning editorial stance, and it is often considered to have a partisan bias in its reporting."
        }
    }
}
\end{minted}
\caption{Sample response for the prompt used in  \S\ \ref{DFR}.}
\label{response_3}
\end{listing*}

\section{Systematic Prompts}

\subsection{Systematic Guidelines}
\label{A:LRDEFS}
Left- and right-winged definitions of the 16 topics as described in  \S\ \ref{App2} can be seen in Listing \ref{app2_defs}.

\begin{listing*}
\begin{minted}[
frame=topline,
framesep=2mm,
baselinestretch=1.1,
fontsize=\footnotesize, 
mathescape, breaklines]{json}
{
    "General Philosophy":{
        "left": "Collectivism: Community over the individual. Equality, environmental protection, expanded educational opportunities, social safety nets for those who need them.",
        "right": "Individualism: Individual over the community. Limited Government with Individual freedom and personal property rights. Competition.",
    },
    "Abortion":{
        "left": "Legal in most cases.",
        "right": "Generally illegal with some exceptions.",
    },
    "Economic Policy":{
        "left": "Income equality; higher tax rates on the wealthy; government spending on social programs and infrastructure; stronger regulations on business. Minimum wages and some redistribution of wealth.",
        "right": "Lower taxes; less regulation on businesses; reduced government spending.  The government should tax less and spend less. Charity over social safety nets. Wages should be set by the free market.",
    },
    "Education Policy":{
        "left": "Favor expanded free, public education. Reduced cost or free college.",
        "right": "Supports homeschooling and private schools. Generally not opposed to public education, but critical of what is taught.",
    },
    "Environmental Policy":{
        "left": "Regulations to protect the environment. Climate change is human-influenced and immediate action is needed to slow it.",
        "right": "Considers the economic impact of environmental regulation. Believe the free market will find its own solution to environmental problems, including climate change. Some deny climate change is human-influenced.",
    },
    "Gay Rights":{
        "left": "Generally support gay marriage; support anti-discrimination laws to protect LGBT against workplace discrimination.",
        "right": "Generally opposed to gay marriage; opposed to certain anti-discrimination laws because they believe such laws conflict with certain religious beliefs and restrict freedom of religion.",
    },
    "Gun Rights":{
        "left": "Favors laws such as background checks or waiting periods before buying a gun; banning certain high capacity weapons to prevent mass shootings.",
        "right": "Strong supporters of the Second Amendment (the right to bear arms), believing it’s a deterrent against authoritarian rule and the right to protect oneself. Generally, does not support banning any type of weaponry.",
    },
    "Health Care":{
        "left": "Most support universal healthcare; strong support of government involvement in healthcare, including Medicare and Medicaid. Generally, support the Affordable Care Act. Many believe healthcare is a human right.",
        "right": "Believe private companies can provide healthcare services more efficiently than government-run programs. Oppose the Affordable Care Act. Insurance companies can choose what to cover and compete with each other. Healthcare is not a right.",
    },
    \end{minted}
    % \caption{Left- and right-winged definitions of the 16 topics as described in Section \ref{App2}.}
    \end{listing*}
    \begin{listing*}
    \begin{minted}[
    frame=bottomline,
    framesep=2mm,
    baselinestretch=1.1,
    fontsize=\footnotesize, 
    mathescape, breaklines]{json}
    {
    "Immigration":{
        "left": "Generally, support a moratorium on deporting or offering a pathway to citizenship to certain undocumented immigrants. e.g., those with no criminal record have lived in the U.S. for 5+ years. Less restrictive legal immigration.",
        "right": "Generally against amnesty for any undocumented immigrants. Oppose a moratorium on deporting certain workers. Funding for stronger enforcement actions at the border (security, wall). More restrictive legal immigration.",
    },
    "Military":{
        "left": "Decreased Spending",
        "right": "Increased Spending",
    },
    "Personal Responsibility":{
        "left": "Strong government to provide a structure. Laws are enacted to protect every individual for an equal society. Safety nets for those in need.",
        "right": "Personal responsibility and it is the government’s role to hold them accountable. Fair competition over safety nets.",
    },
    "Regulation":{
        "left": "Government regulations are needed to protect consumers and the environment.",
        "right": "Government regulations hinder free-market capitalism and job growth.",
    },
    "Social Views":{
        "left": "Based on community and social responsibility. Gay couples to get equal rights like everyone else (e.g. marriage); abortion should be legal; support embryonic stem cell research. Support restrictions and regulations around the right to bear arms. Supports taxpayer funding of Planned Parenthood.",
        "right": "Based on individual rights and justice. Opposed to gay marriage, abortion, and embryonic stem cell research. Support the right to bear arms, death penalty, and personal responsibility as an individual. Opposed to taxpayer funding of Planned Parenthood.",
    },
    "Taxes":{
        "left": "Progressive taxation (high-income earners taxed at a higher rate). Generally not opposed to raising taxes to fund the government.",
        "right": "Tend to favor a “flat tax” (same tax rate regardless of income). Generally opposed to raising taxes to fund the government.",
    },
    "Voter ID":{
        "left": "Against voter ID laws citing an undue burden on lower-income groups causing them to be disenfranchised, and that there is virtually no evidence of voter fraud actually occurring.",
        "right": "For voter identification laws to combat alleged voter fraud.",
    },
    "Business Rights":{
        "left": "Supports unions and worker protections. Raising the minimum wage. Higher corporate taxes.",
        "right": "Favors business owners and corporations with the expectation higher profits will result in higher wages through a free-market. Generally opposed to a minimum wage. Lower corporate taxes.",
    }
}
\end{minted}
\caption{Left- and right-winged definitions of the 16 topics as described in  \S\ \ref{App2}.}
\label{app2_defs}
\end{listing*}

\subsection{Response to Systematic Prompt}
\label{A:AppRes}

A sample response for the prompt used in  \S\ \ref{App2} can be seen in Listing \ref{sys_res}.

\begin{listing*}
\begin{minted}[
frame=topline,
framesep=2mm,
baselinestretch=1.1,
fontsize=\footnotesize, 
mathescape, breaklines]{json}
{
    "foxnews.com": {
        "General Philosophy": {
            "leaning": "right",
            "reason": "Fox News is known for its conservative and right-leaning perspective on political and social issues, promoting individualism and limited government."
        },
        "Abortion": {
            "leaning": "right",
            "reason": "Fox News is known for its conservative and right-leaning perspective on social issues, including abortion. It generally aligns with the right-leaning stance of being generally illegal with some exceptions."
        },
        "Economic Policy": {
            "leaning": "right",
            "reason": "Fox News is known for its conservative and right-leaning perspective on economic policy, advocating for lower taxes, less regulation on businesses, and reduced government spending."
        },
        "Education Policy": {
            "leaning": "right",
            "reason": "Fox News is known for its conservative and right-leaning perspective, often supporting homeschooling and private schools while being critical of public education policies."
        },
        "Environmental Policy": {
            "leaning": "right",
            "reason": "Fox News is known for its conservative and right-leaning perspective on environmental policy, often emphasizing the economic impact of regulations and expressing skepticism about human-influenced climate change."
        },
        "Gay Rights": {
            "leaning": "right",
            "reason": "Fox News is generally opposed to gay marriage and may be opposed to certain anti-discrimination laws due to religious beliefs."
        },
        "Gun Rights": {
            "leaning": "right",
            "reason": "Fox News is known for its conservative and right-leaning perspective on various issues, including gun rights. They generally support the Second Amendment and the right to bear arms without significant restrictions."
        },
        "Health Care": {
            "leaning": "right",
            "reason": "Fox News is known for its conservative and right-leaning perspective on healthcare, often opposing government involvement and supporting private companies in providing healthcare services."
        },
        "Immigration": {
            "leaning": "right",
            "reason": "Fox News is known for its conservative and right-leaning perspective on immigration, often advocating for stronger enforcement actions at the border and opposing amnesty for undocumented immigrants."
        },
        "Military": {
            "leaning": "right",
            "reason": "Fox News is known for its conservative and right-leaning perspective, often advocating for increased military spending and a strong military presence."
        },
        
    \end{minted}
    \end{listing*}
    \begin{listing*}
    \begin{minted}[
    frame=bottomline,
    framesep=2mm,
    baselinestretch=1.1,
    fontsize=\footnotesize, 
    mathescape, breaklines]{json}
    {

        "Personal Responsibility": {
            "leaning": "right",
            "reason": "Fox News is known for its conservative and right-leaning perspective, often emphasizing personal responsibility and limited government intervention."
        },
        
        "Regulation": {
            "leaning": "right",
            "reason": "Fox News is known for its conservative and right-leaning perspective, which often opposes government regulations as hindering free-market capitalism and job growth."
        },
        "Social Views": {
            "leaning": "right",
            "reason": "Fox News is known for its conservative and right-leaning perspective on social issues, including opposition to gay marriage, abortion, and embryonic stem cell research, as well as support for the right to bear arms and opposition to taxpayer funding of Planned Parenthood."
        },
        "Taxes": {
            "leaning": "right",
            "reason": "Fox News tends to favor a 'flat tax' and is generally opposed to raising taxes to fund the government."
        },
        "Voter ID": {
            "leaning": "right",
            "reason": "Fox News is known for its conservative and right-leaning perspective, and it generally supports voter identification laws to combat alleged voter fraud."
        },
        "Business Rights": {
            "leaning": "right",
            "reason": "Fox News is known for its conservative and right-leaning perspective, often favoring business owners and corporations with the expectation of higher profits resulting in higher wages through a free-market approach."
        }
    }
}

\end{minted}
\caption{A sample response for the prompt used in  \S\ \ref{App2}.}
\label{sys_res}
\end{listing*}

\section{Zero-Shot Predictions}
\subsection{LLM Response Given a Media Name}
\label{A:LLMFailedResponse}
A response where LLM was unable to predict the factuality of reporting:

\begin{minted}[
frame=lines,
framesep=1mm,
baselinestretch=1.1,
fontsize=\footnotesize,
mathescape, 
breaklines]{json}
{
    "input": "vancouvertimes.org",
    "output": -1
}
\end{minted}

\subsection{Political Bias Prompt Given a Media Name}
\label{A:PromptBiasGivenMediaName}

The following is the prompt used to predict the political bias of the media given its name:

\begin{minted}[
frame=lines,
framesep=1mm,
baselinestretch=1.1,
fontsize=\footnotesize,
mathescape, 
breaklines]{python}
system_prompt = '''You are an expert in media analysis. Classify the bias of a media given its name ONLY from one of three categories from the list provided below:
- left
- center
- right
Return -1 if you can not classify.'''   
user_prompt = "{domain}"
\end{minted}

\subsection{Summarization}
\label{A:Summ}
The following prompt in Listing \ref{prompt:sum} is used to summarize the articles from the news media as described in  \S\ \ref{CSGA}, where the \textbf{\{article\}} placeholder is replaced with the article from the news media.

\begin{listing*}
\begin{minted}[
frame=lines,
framesep=2mm,
baselinestretch=1.1,
fontsize=\footnotesize, 
mathescape, breaklines]{python}
system_prompt = '''Summarize the following news article in 250-300 words. Ensure the summary covers the article's key points and main details.'''   
user_prompt = "{article}"
\end{minted}
\caption{Prompt used to summarize the articles from news media as described in  \S\ \ref{CSGA}.}
\label{prompt:sum}
\end{listing*}

\subsection{Political Bias Prompt Given Media Articles}

The following is the prompt used to predict the political bias of the media given its summarized articles:

\label{A:PromptBiasGivenMediaArticle}
\begin{minted}[
frame=lines,
framesep=1mm,
baselinestretch=1.1,
fontsize=\footnotesize,
mathescape, 
breaklines]{python}
system_prompt = '''You are an expert in media analysis. Classify the bias of the given news article of {media} into one of three categories from the list provided below:
- left
- center
- right
Return -1 if you can not classify.'''   
user_prompt = "{article}"
\end{minted}

\section{Dataset}
\label{A:Dataset}

\begin{figure*}
    \centering
    \includegraphics[width=0.9\linewidth]{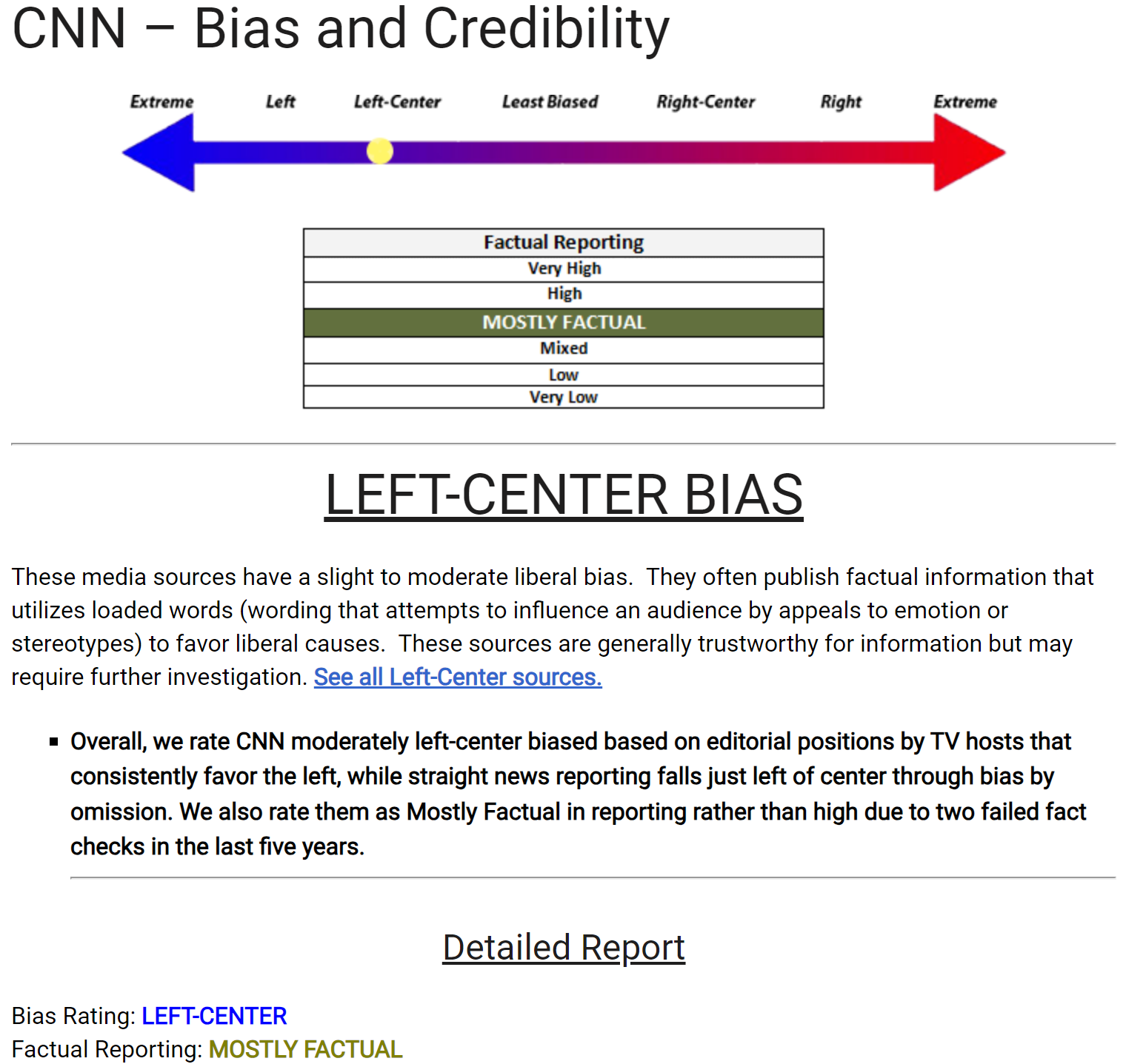}
    \caption{An example of annotation of a news outlet from MBFC. Source:  \href{https://mediabiasfactcheck.com/left/cnn-bias/} {www.mediabiasfactcheck.com}.}
    \label{fig:ExampleAnnotation}
\end{figure*}

An example annotation for cnn.com provided by MBFC is shown in Figure \ref{fig:ExampleAnnotation}. Detailed dataset statistics are shown in Table~\ref{tab:Dataset}

% \begin{table}[t]
%     \caption{Dataset}
%     \label{tab:Dataset}
%     \begin{center}
%     %\vspace{-2em}
%     \setlength{\tabcolsep}{4pt}
%     \scalebox{0.8}{%
%     \begin{tabular}{lr lr | lr lr}
%     \toprule
%     \multicolumn{4}{c}{\textbf{ACL-2020}} & \multicolumn{4}{c}{\textbf{\textcolor{red}{EMNLP-2024}}} \\
%      \midrule
%     \multicolumn{2}{c}{Political Bias} & \multicolumn{2}{c}{Factuality} & \multicolumn{2}{c}{Political Bias} & \multicolumn{2}{c}{Factuality} \\ 
%      \midrule
%     Left & 243 & Low & 542  & Left & 398 & Low & 597 \\
%     Center & 272 & Mixed & 268  & Center & 913 & Mixed & 1200 \\
%     Right & 349 & High & 349  & Right & 831 & High & 2395 \\    
%     \bottomrule
%     \end{tabular}
%     }
%     \end{center}
% \end{table}

\begin{table}[!t]
    \begin{center}
    %\vspace{-2em}
    \setlength{\tabcolsep}{4pt}
    \scalebox{0.9}{%
    \begin{tabular}{lr | lr}
    \toprule
    \multicolumn{4}{c}{\textbf{MBFC Dataset}} \\
     \midrule
    \multicolumn{2}{c}{Political Bias} & \multicolumn{2}{c}{Factuality} \\ 
     \midrule
    Left & 398 & Low & 597 \\
    Left-Center & 600  & Mixed & 1200\\
    Center & 913 & High & 2395 \\
    Right-Center & 907 &  &  \\
    Right & 831 &  &  \\  
    \midrule
    \textbf{Total} & \textbf{3649} & \textbf{4192} & \\
    \bottomrule
    \end{tabular}
    }
    \caption{Distribution of labels in our dataset.}
    \label{tab:Dataset}
    \end{center}
\end{table}

\end{document}